\documentclass[10pt,twocolumn,letterpaper]{article}

\usepackage{cvpr}              
\usepackage{float}

\definecolor{cvprblue}{rgb}{0.21,0.49,0.74}
\usepackage[pagebackref,breaklinks,colorlinks,allcolors=cvprblue]{hyperref}
\usepackage{bm}
\usepackage{graphicx}
\usepackage{amsmath}
\usepackage{amssymb}
\usepackage{booktabs}
\usepackage{float}
\usepackage{wrapfig}
\usepackage{multirow}

\title{F3D-Gaus: Feed-forward 3D-aware Generation on ImageNet \\with Cycle-Aggregative Gaussian Splatting}

\author{Yuxin Wang\\
HKUST\\
{\tt\small ywangom@connect.ust.hk}
\and
Qianyi Wu\\
Monash University\\
{\tt\small qianyi.wu@monash.edu}
\and
Dan Xu$^*$\\
HKUST\\
{\tt\small danxu@cse.ust.hk}
}

\begin{document}
\maketitle
\begin{abstract}
This paper tackles the problem of generalizable 3D-aware generation from monocular datasets, {e.g.}, ImageNet~\cite{deng2009imagenet}. The key challenge of this task is learning a robust 3D-aware representation without multi-view or dynamic data, while ensuring consistent texture and geometry across different viewpoints. 
Although some baseline methods are capable of 3D-aware generation, the quality of the generated images still lags behind state-of-the-art 2D generation approaches, which excel in producing high-quality, detailed images.  
To address this severe limitation, we propose a novel feed-forward pipeline based on pixel-aligned Gaussian Splatting, coined as~\textbf{F3D-Gaus}, which can produce more realistic and reliable 3D renderings from monocular inputs. 
In addition, we introduce a self-supervised cycle-aggregative constraint to enforce cross-view consistency in the learned 3D representation. This training strategy naturally allows aggregation of multiple aligned Gaussian primitives and significantly alleviates the interpolation limitations inherent in single-view pixel-aligned Gaussian Splatting. 
Furthermore, we incorporate video model priors to perform geometry-aware refinement, enhancing the generation of fine details in wide-viewpoint scenarios and improving the model’s capability to capture intricate 3D textures. 
Extensive experiments demonstrate that our approach not only achieves high-quality, multi-view consistent 3D-aware generation from monocular datasets, but also significantly improves training and inference efficiency. Project Page: \url{https://w-ted.github.io/publications/F3D-Gaus} 
\end{abstract}

\vspace{-8pt}
\section{Introduction}
\label{sec:intro}

Generating 3D-aware content from a single image has numerous applications in augmented reality and gaming, enhancing immersive experiences. However, achieving 3D awareness typically requires substantial supervision. Recent 3D-aware generation methods rely on optimization based on 2D/3D diffusion priors~\cite{tang2023make, you2024nvs}, multi-view image supervision~\cite{szymanowicz2024splatter, tang2025lgm, hong2023lrm, zhang2025gs, wang2023imagedream, shi2023MVDream, zeronvs}, or fine-tuning on video data~\cite{gao2024cat3d, yu2024viewcrafter}. Among these, inference-time optimization approaches~\cite{gao2024cat3d, tang2023make, you2024nvs} often suffer from inefficiency, making them impractical for real-world applications, particularly in time-sensitive scenarios or resource-constrained environments. Consequently, there is a growing demand for feed-forward models capable of directly inferring 3D structure from single images. 

\begin{figure}[!tp]
  \centering   \includegraphics[width=1.0\linewidth]{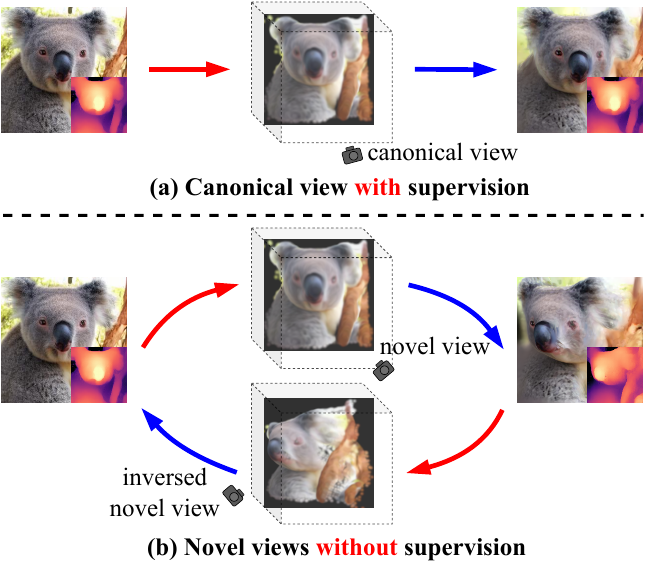}
   \vspace{-17pt}\caption{\textbf{Illustration of our motivation for cycle self-supervised training. } For monocular datasets: (a) supervision is naturally available for the canonical view. (b) For novel views, where supervision is absent, we use the rendered novel-view image as input to obtain its 3D representation. This 3D representation is then re-rendered from the canonical view, where supervision is available. \textcolor{red}{Red arrows} indicate feed-forward 3D representation prediction from a monocular image, while \textcolor{blue}{blue arrows} represent the rendering processes from 3D representations at different specific viewpoints. }
   \label{fig:teaser}
   \vspace{-16pt}
\end{figure}

While most successful attempts~\cite{hong2023lrm, szymanowicz2024splatter, zeronvs, yu2024viewcrafter} rely on multi-view or video data, such datasets are often difficult and costly to obtain. In contrast, monocular data sets, consisting of single-frame images, are significantly more abundant and easier to collect. They can be sourced from the web or captured with consumer-grade cameras without requiring specialized equipment or controlled environments. Moreover, the scalability of monocular datasets enables training on large-scale data, which can enhance generalization across diverse scenes and object categories. These considerations motivate us to explore a key research question: \emph{Can we design a feed-forward 3D-aware generation framework that is trained purely on monocular image datasets?}

This problem has long been central to computer vision, closely related to 3D representation learning. Traditional methods for 3D-aware generation from a single image typically employ explicit 3D representations, such as voxels~\cite{choy20163d}, point clouds~\cite{Wei2021CVPR}, or meshes~\cite{wang2018pixel2mesh, gkioxari2019mesh,wu2020unsupervised}. However, these approaches suffer from limitations in appearance modeling and often struggle with generalization beyond specific object categories. More recently, radiance field-based representations have revolutionized 3D-aware generation due to their ability to model both geometry and appearance in a unified manner. The advent of NeRF~\cite{mildenhall2021nerf} has led to significant progress in learning 3D structures from monocular datasets~\cite{chan2022efficient, 3dgp,reddy2024g3dr}. However, the implicit nature of NeRF, where the entire 3D asset is represented by triplane-based feature encoding and a shared MLP decoder, limits rendering quality. A promising alternative is the recently emerging 3D Gaussian Splatting (3DGS)~\cite{3dgs}, which represents radiance fields using discrete Gaussian primitives. This formulation improves rendering efficiency and inspires a novel feedforward framework: pixel-aligned Gaussian Splatting, designed to infer 3D-aware representations in a single forward pass using only monocular training data.

Despite its advantages, training on monocular datasets presents a fundamental challenge: since supervision is available only for frontal views, generating plausible novel viewpoints is non-trivial. Previous monocular-based methods address this issue using discriminators~\cite{3dgp}, semantic/geometry constraints~\cite{reddy2024g3dr}, or iterative in-painting techniques~\cite{xiang2023ivid}. While pixel-aligned GS offers advantages in rendering realism, it struggles with significant viewpoint changes. The occluded regions in the original view are difficult to interpolate, leading to incomplete geometry and artifacts in novel views.

To tackle this challenge, we introduce a cycle-aggregative strategy. It ensures that 3D representations from novel viewpoints are both aligned with and complementary to those from the original viewpoint. By enforcing this cycle consistency through aggregation, the model naturally learns to extrapolate across views by fusing multiple representations, enhancing 3D coherence during inference. Additionally, we incorporate a video in-painting model~\cite{zhou2023propainter} guided by geometry information to correct inconsistencies in geometry and texture caused by large viewpoint shifts, resulting in more reliable and visually consistent outputs.

Through extensive evaluation, our approach achieves state-of-the-art realism in 3D rendering, surpassing existing benchmarks. We validate its robustness across diverse datasets to demonstrate strong generalization. Furthermore, our method significantly improves computational efficiency, reducing both training and inference times while maintaining high-quality outputs.

In summary, our work presents a comprehensive solution for realistic and efficient 3D-aware generation from monocular datasets, marking a significant advancement in 3D content generation. Our contributions are threefold: 

\begin{itemize} 
\item We pioneer 3D-aware generation from single image using a generalizable, feed-forward Gaussian Splatting representation, achieving both high efficiency and superior rendering quality on monocular datasets.
\item We introduce a self-supervised cycle-aggregative training strategy to enhance the capability of pixel-aligned Gaussian Splatting, enabling more robust novel view synthesis from monocular images.
\item We mitigate artifacts caused by large viewpoint shifts by integrating geometry-aware video priors, further refining 3D-aware representations.
\end{itemize}
\vspace{-3pt}
\section{Related Works}
\label{sec:related_works}

\begin{figure*}[!tp]
  \includegraphics[width=0.98\linewidth]{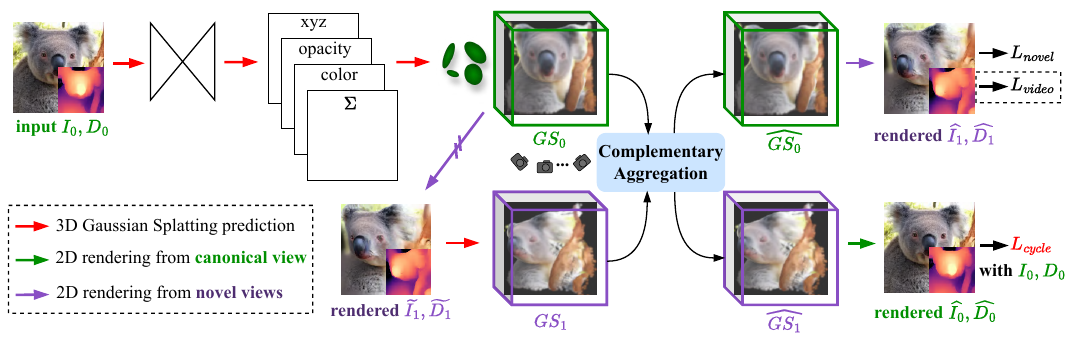}
  \vspace{-14pt}
    \caption{\textbf{Illustration of our overall framework.} Given a single RGB image $I_0$ and depth map $D_0$, our model directly feeds them forward to output the pixel-aligned Gaussian Splatting representation $GS_0$, which can be used for novel view synthesis. After obtaining the 3DGS representation, we render the image $\tilde{I_1}$ and depth maps $\tilde{D_1}$ for the novel view, and then output its corresponding 3DGS $GS_1$. These two 3DGS representations are subsequently aggregated to produce the images for supervision. This novel self-supervised training strategy enforces cycle-aggregative 3D representation learning across different views, allowing the generalized 3DGS representations to reinforce each other, thereby collaboratively enhancing the overall 3D representation capability. }
  \label{fig:pipeline}
  \vspace{-16pt}
\end{figure*}

\subsection{Novel view synthesis for single input}
\noindent\textbf{With multi-view dataset.} Novel view synthesis from a single image can be achieved through various methods. One straightforward strategy is to train or fine-tune models using multi-view or video data with camera information, which data naturally provides strong multi-view cues for 3D-aware generation. 
Some methods train models from scratch using multi-view data~\cite{hong2023lrm, zeronvs, szymanowicz2024splatter, tang2025lgm, li2022_infinite_nature_zero, zhong2024cvt}, while others fine-tune pre-trained models for image~\cite{wang2023imagedream, shi2023MVDream, gao2024cat3d, hollein2024viewdiff} or video generation~\cite{guo2023animatediff, wang2024motionctrl, melaskyriazi2024im3d, voleti2024sv3d, yu2024viewcrafter} to enable camera motion control. 
Another category of methods optimizes scene-specific representations using 2D/3D diffusion priors~\cite{tang2023make, Magic123, hollein2023text2room, you2024nvs}. However, these approaches lack generalization ability, as they are tailored to individual objects or scenes. 

\noindent\textbf{With only monocular dataset. }
The most similar approach to ours trains a generalized model solely on monocular datasets for feed-forward predictions~\cite{chan2022efficient,3dgp,xiang2023ivid,sargent2023vq3d,reddy2024g3dr}. 
3DGP~\cite{3dgp} pioneered 3D-aware generation on ImageNet using a 3D-aware GAN for RGB-D prediction, with a discriminator for realism and a depth-adapter for geometry refinement. 
G3DR~\cite{reddy2024g3dr} employs a generalized tri-plane representation with semantic and perceptual losses for novel view rendering. 
Our method operates in a similar setting, but combines pixel-aligned Gaussian Splatting with a cycle-aggregative training, achieving more realistic renderings than tri-plane approaches. 
IVID~\cite{xiang2023ivid} uses 2D diffusion for iterative RGB-D in-painting, but struggles with novel object categories, as it relies on ImageNet class labels.
Real3D~\cite{jiang2024real3d} uses a self-supervised strategy for 3D asset generation. 
However, our approach differs in both setting and technical design: We introduce a generalizable feed-forward GS for monocular datasets, while Real3D relies on LRM~\cite{hong2023lrm} and combines monocular and multi-view data. 
We additionally propose a complementary aggregation design for pixel-aligned GS and integrate video priors to enhance the learning process.

\subsection{3D Gaussian Splatting}
Recently, 3D Gaussian Splatting (3DGS)~\cite{3dgs} stands as a leading representation for novel view synthesis, enabling real-time rendering with state-of-the-art visual quality. It models the scene as learnable 3D Gaussian primitives with 3D coordinates, opacity, anisotropic covariance, and color features. 
Recent works extend 3DGS to tasks like scene understanding~\cite{cao20243dgs,zhang2024egogaussian,shi2024language,zhou2024hugs,zhou2024feature}, editing~\cite{ye2025gaussian,wang2024gaussianeditor,chen2024gaussianeditor,wang2025learning,liu2024infusion}, and surface reconstruction~\cite{lyu20243dgsr,wang2024pygs,chen2024pgsr,Wu2024gsrec,huang20242d,yan2024gs,Yu2024GOF}. 
However, 3DGS was originally designed for scene-specific novel view synthesis. To generalize 3DGS prediction from single or multiple images, recent works~\cite{szymanowicz2024splatter, tang2025lgm, szymanowicz2024flash3d, charatan2024pixelsplat, chen2025mvsplat, chen2024mvsplat360, smart2024splatt3r} have emerged. Splatter-Image~\cite{szymanowicz2024splatter} pioneered pixel-aligned 3DGS prediction from RGB images, enabling fast feed-forward training and inference. LGM~\cite{tang2025lgm} combines pre-trained multi-view image generation models with an image-to-3DGS model for object-level 3DGS inference from a single image. Flash3D~\cite{szymanowicz2024flash3d} integrates depth estimation for scene-level 3DGS prediction from a single image. 
While these methods allow feed-forward inference from a single image to 3DGS, they still rely on multi-view data for training or use multi-view image generation models as priors. In contrast, our approach focuses on generalized 3DGS prediction without requiring multi-view images for training. 
\section{The Proposed Framework: F3D-Gaus}
\label{sec:method}

Given a dataset of images and their corresponding monocular depth maps, represented as $\left\{(I_i, D_i)| i=0, \dots, N\right\}$, we aim to train a generalized model that takes a single RGB-D input and outputs the associated 3DGS representation in a feed-forward manner. This representation can then be utilized for novel view synthesis. The monocular depth maps $D_i$ can be easily obtained using monocular depth estimation models~\cite{Wei2021CVPR, ke2023repurposing, depth_anything_v1}. 

As shown in Fig.~\ref{fig:pipeline}, we use a U-Net-based generalized 3DGS model as the backbone. Given an input image $I_0$ and its depth map $D_0$, the model predicts a pixel-aligned 3DGS representation, which can render images and depth maps from arbitrary viewpoints. The details of the generalized 3DGS model are provided in Sec.~\ref{prelimilnary}. 
As shown in Fig.~\ref{fig:teaser}, the rendered outputs can be supervised in the canonical view but the monocular dataset lacks supervision for novel views. We propose a cycle-aggregative self-supervised training strategy in Sec.~\ref{sec:cycle} to address this. 
Additionally, our geometry-guided refinement method introduced in Sec.~\ref{sec:geo} further identifies and refines artifacts.

\subsection{Preliminary}
\label{prelimilnary}
\noindent\textbf{3D Gaussian Splatting.}
Each Gaussian primitive is characterized by its 3D coordinates $\bm{\mu}$, color features $\bm{c}$, opacity, scale matrix $S$, and rotation matrix $R$. 
With these attributes, the Gaussians are defined by the covariance matrix $\Sigma = RSS^TR^T$ centered at point $\bm{\mu}$ following $G(\bm{x}) = \exp^{-\frac{1}{2}(\bm{x}-\bm{\mu})^T\Sigma^{-1}(\bm{x}-\bm{\mu})}$.
The covariance matrix is projected onto the 2D plane following~\cite{zwicker2001ewa}, allowing us to compute the projected Gaussian and apply alpha-blending to obtain the final color on the image plane:
\begin{equation}
\setlength{\abovedisplayskip}{3pt}
\setlength{\belowdisplayskip}{6pt}
    \hat{I} = \sum_{k=1}^K \bm{c}_k\alpha_k\Pi_{j=1}^{k-1}(1-\alpha_j),
    \label{equ:2}
\end{equation}
where $K$ is the number of sampling points along the ray and $\alpha$ is derived from the projected Gaussian of $G(\bm{x})$ and its corresponding opacity. 
More details can be found in~\cite{3dgs}.

\noindent\textbf{Pixel-aligned 3DGS.} 
Splatter-Image~\cite{szymanowicz2024splatter} introduces a framework for predicting pixel-aligned 3DGS from images. Given an input image $I_0$, it predicts Gaussian attribute maps, including 3D coordinates, opacity, color features, scale, and rotation, matching the same input spatial size with $n = H \times W$ Gaussian primitives. Scale and rotation define the covariance matrix. Together with other attributes, images and depth maps can be rendered from arbitrary viewpoints. 
We use a similar U-Net model as the backbone, while 3DGS's coordinates are computed by adding the input depth $D_0$ to the predicted offset.

\subsection{Cycle-aggregative Self-supervised Strategy}
\label{sec:cycle}
The lack of novel view supervision motivates us to propose a cycle-aggregative self-supervised strategy in this section. 
It aims to ensure the multi-view consistency of the predicted 3DGS. It has two key components: complementary aggregation and cycle supervision. The details are given below.

\noindent\textbf{Canonical view reconstruction.}
We assume each image in the monocular dataset is captured from a canonical view. During training, we pass RGB-D input ($I_0$, $D_0$) through the model to obtain the output $GS_0$. We then randomly sample camera views to render the scene from either the canonical view, represented as $\text{view}_0$, or a randomly selected novel view $\text{view}_1$. If the canonical view is rendered, as shown in Fig.~\ref{fig:teaser} (a), a reconstruction loss is applied directly to the predicted image $\hat{I_0}$ and depth map $\hat{D_0}$, effectively performing RGB-D image reconstruction. 
\begin{equation}
\setlength{\abovedisplayskip}{8pt}
\setlength{\belowdisplayskip}{8pt}
    \mathcal{L}_{\text{recon}} = \lVert \hat{I_0} - I_0 \rVert_1 + \lVert \hat{D_0} - D_0 \rVert_1.
    \label{equ:1}
\end{equation}
The backpropagated gradients update the parameters of our U-Net backbone via the 3DGS attribute maps.

\begin{figure}[!tp]
  \centering
  \includegraphics[width=1.0\linewidth]{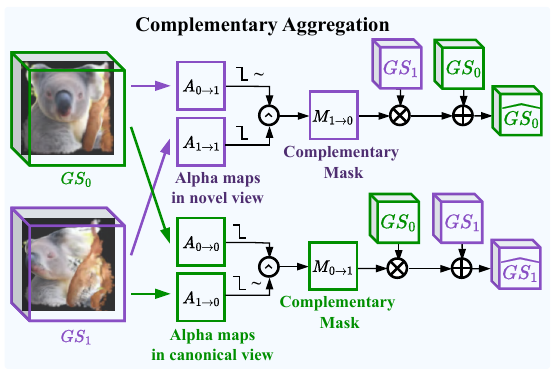}
  \vspace{-20pt}
  \caption{\textbf{Illustration of the proposed cycle-aggregative self-supervised strategy.} We guide complementary aggregation by leveraging the differences between the alpha maps of the two 3DGS from different viewpoints. }
  \label{fig:comp}
  \vspace{-20pt}
\end{figure}

\noindent\textbf{Complementary aggregation.}
The idea behind complementary aggregation is as follows: our method can predict the canonical $GS_0$ from the input $I_0$ and $D_0$, we assume that if we had a side-view image of the same object (even though the dataset lacks ground truth for side views, we can render one from $GS_0$, denoted as $\tilde{I_1}$ and $\tilde{D_1}$ in Fig.~\ref{fig:pipeline}), we could also obtain a novel view $GS_1$. We aim to enforce multi-view consistency between the rendered $\tilde{I_1}$, $\tilde{D_1}$ and the original $I_0$, $D_0$, which implies that $GS_0$ and $GS_1$ should also be multi-view consistent. Another intuition is that, $GS_0$ and $GS_1$ are 3D representations from two different views but they represent the same scene, they should also be complementary. 
Thus, we designed complementary aggregation to ensure that $GS_0$ and $GS_1$ are both multi-view consistent and complementary.

As shown in Fig.~\ref{fig:comp}, after obtaining $GS_0$ and $GS_1$, we apply an aggregation strategy to update both 3DGSs. Specifically, we first render the alpha maps $\hat{A}$ for both the $\text{view}_0$ and the $\text{view}_1$ from the two 3DGS representations. 
\begin{equation}
\setlength{\abovedisplayskip}{6pt}
\setlength{\belowdisplayskip}{6pt}
    \hat{A} = \sum_{k=1}^K \alpha_k\Pi_{j=1}^{k-1}(1-\alpha_j),
\end{equation}
$\alpha$ and $K$ have the same meaning as Equ.~\ref{equ:2}. In Fig.~\ref{fig:comp}, $A_{i \rightarrow j}$ represents the alpha map rendered from $GS_i$ in $\text{view}_j$. 
We binarize ($\urcorner \hspace{-0.15em} \llcorner$) the alpha maps to detect invisible region in $GS_i$ at specific locations in $\text{view}_j$. Using logical operations, we then generate two complementary masks:
\begin{equation}
\setlength{\abovedisplayskip}{6pt}
\setlength{\belowdisplayskip}{6pt}
\begin{split}
    M_{1 \rightarrow 0} = \neg (A_{0 \rightarrow 1} < \tau) \land (A_{1 \rightarrow 1} < \tau) \\
    M_{0 \rightarrow 1} = (A_{0\rightarrow 0} < \tau) \land \neg (A_{1 \rightarrow 1} < \tau)
\end{split}
\end{equation}
We set $\tau = 0.5$, where $\neg$ represents logical negation, and $\land$ represents the logical and. $M_{1 \rightarrow 0}$ identifies the locations in $\text{view}_1$ where $GS_0$ has holes but $GS_1$ has valid values, indicating the specific primitives in $GS_1$ that can contribute to $GS_0$. These complementary masks are then used to update $GS_0$ and $GS_1$, as shown in Fig.~\ref{fig:comp}.
\begin{equation}
\setlength{\abovedisplayskip}{3pt}
\setlength{\belowdisplayskip}{3pt}
\begin{split}
    \hat{GS_0} &= \textbf{Concat}(GS_0, GS_1[M_{1 \rightarrow 0}]) \\
    \hat{GS_1} &= \textbf{Concat}(GS_1, GS_0[M_{0 \rightarrow 1}])
\end{split}
\end{equation}
In the equations above, $[M_{i \rightarrow j}]$ uses the complementary mask to select the corresponding pixel-aligned primitives. The function $\textbf{Concat}(*, *)$ combines two sets of 3DGS primitives into a new set. 

\noindent\textbf{Cycle supervision.}
As shown in Fig.~\ref{fig:pipeline}, after obtaining the updated $\hat{GS_0}$ and $\hat{GS_1}$, we render $\hat{I_1}$, $\hat{D_1}$ from $\text{view}_1$ and $\hat{I_0}$, $\hat{D_0}$ from $\text{view}_0$, respectively. Note that the rendered views are opposite to the source input view of the GS representations. This augmentation is designed to maximize the use of supervision. 
For $\hat{I_0}$ and $\hat{D_0}$, we use the original $I_0$ and $D_0$ for supervision, referred to as cycle consistency loss: 
\begin{equation}
\setlength{\abovedisplayskip}{6pt}
\setlength{\belowdisplayskip}{6pt}
    \mathcal{L}_{\text{cycle}} = \lVert \hat{I_0} - I_0 \rVert_1 + \lVert \hat{D_0} - D_0 \rVert_1
    \label{equ:9}
\end{equation}
The cycle loss $\mathcal{L}_{\text{cycle}}$ in Equ.~\ref{equ:9} looks the same as the reconstruction loss in Equ.~\ref{equ:1}, but they correspond to sampled canonical and novel views, respectively. The predictions of $\hat{I}_0$ and $\hat{D}_0$ differ in each case. 
Due to our complementary aggregation design, the gradients from $\mathcal{L}_{\text{cycle}}$ backpropagate through both $GS_0$ and $GS_1$. We block gradient propagating to $GS_0$ via $\tilde{I}_1$ and $\tilde{D}_1$, as we found that allowing it hinders the learning process.

\noindent\textbf{Overall losses.} 
For $\hat{I_1}$ and $\hat{D_1}$, we apply perceptual loss~\cite{johnson2016perceptual} and CLIP losses~\cite{radford2021learning} following~\cite{reddy2024g3dr}. In addition, we use photometric loss to the novel views by warping textures from the frontal view with depth $D_0$. 
\begin{equation}
\setlength{\abovedisplayskip}{6pt}
\setlength{\belowdisplayskip}{6pt}
\begin{aligned}
    \mathcal{L}_{\text{perp}} &= \sum_{l} \left\lVert \phi_l(\hat{I}_1) - \phi_l(I_0) \right\rVert_2 \\
    \mathcal{L}_{\text{CLIP}} &= \left\lVert \phi_{\text{CLIP}}(\hat{I}_1) - \phi_{\text{CLIP}}(I_0) \right\rVert_1 \\
    \mathcal{L}_{\text{photo}} &= \left\lVert I_0(x_i) - \hat{I_1}\left(P(D_0(x_i), \pi, T_{0 \rightarrow 1})\right) \right\rVert_1 
\end{aligned}
\end{equation}
$\phi_l$ represents the feature extractor from the first $l$ layers of a perceptual model, while $\phi_{\text{CLIP}}$ refers to the CLIP feature extractor. $\hat{I}_1(P(D_0(x_i), \pi, T_{0 \rightarrow 1}))$ denotes the RGB value of the reprojected predicted image $\hat{I}_1$ at pixel $x_i$, where $\pi$ is the camera intrinsic matrix, $T_{0 \rightarrow 1}$ is the extrinsic matrix, $P$ is the projection function transforming from $\text{view}_0$ to $\text{view}_1$. We then define the losses for the novel view as follows: 
\begin{equation}
    \setlength{\abovedisplayskip}{6pt}
    \setlength{\belowdisplayskip}{6pt}
    \mathcal{L}_{\text{novel}} = \mathcal{L}_{\text{photo}} + \lambda_{\text{perp}} \mathcal{L}_{\text{perp}} + \lambda_{\text{CLIP}} \mathcal{L}_{\text{CLIP}}
\label{equ:novel}
\end{equation}
Note that novel loss $\mathcal{L}_{\text{novel}}$ defined in Equ.~\ref{equ:novel} is not part of our cycle-aggregative training strategy, but rather a component of our baseline method. Even in the baseline method mentioned in the ablation study in Sec.~\ref{sec:abl}, the novel loss is also applied (to $\tilde{I_1}$ and $\tilde{D_1}$ at that time). 
In addition to the above losses, we apply a total-variation regularization loss to enforce smoothness in the rendered depth. 
\begin{equation}
\setlength{\abovedisplayskip}{6pt}
\setlength{\belowdisplayskip}{6pt}
\mathcal{L}_{\text{reg}} = L_{\text{tv}}(\hat{D})
\end{equation}
The overall training loss is shown below, where $\lambda$ represents the weight of each loss term. 
\begin{equation}
\setlength{\abovedisplayskip}{6pt}
\setlength{\belowdisplayskip}{6pt}
\mathcal{L}_{\text{total}} =
\begin{cases}
\mathcal{L}_{\text{recon}} + \lambda_{\text{reg}} \mathcal{L}_{\text{reg}}, & \small{\text{for canonical view}}, \\
\begin{aligned}
\mathcal{L}_{\text{cycle}} &+ \lambda_{\text{reg}} \mathcal{L}_{\text{reg}} \\
&+ \lambda_{\text{novel}} \mathcal{L}_{\text{novel}}
\end{aligned} & \small{\text{for novel views}}.
\end{cases}
\label{equ:total}
\end{equation}

\begin{figure}[!tp]
  \centering
  \includegraphics[width=1.0\linewidth]{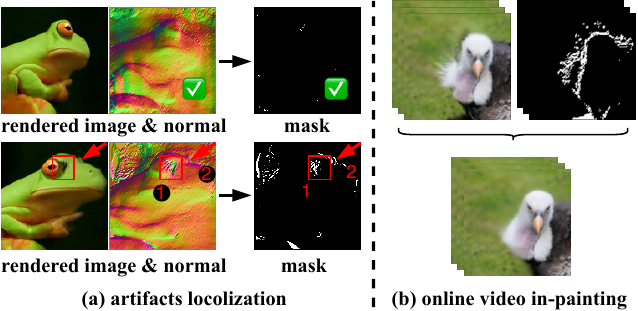}
  \vspace{-20pt}
  \caption{\textbf{Illustration of geometry-guided texture refinement.} (a) illustrates artifact localization in novel views, while (b) shows geometry mask-guided sequence in-painting. }
  \label{fig:normal}
  \vspace{-18pt}
\end{figure}

\noindent\textbf{Inference-time aggregation.} 
During training, we use a cycle-aggregative training strategy by aggregating GS representations from two views and aligning them via cycle supervision. This naturally extends to two-view aggregation during inference, enhancing representational capacity. 
At inference time, given inputs $I_0$ and $D_0$, we first obtain $GS_0$. We then render an image from $GS_0$ from the novel view. Next, we run another inference step to obtain novel-view representation $GS_1$, which are concatenated with $GS_0$ to form the final GS representation.

\subsection{Geometry-guided Texture Refinement}
\label{sec:geo}
We found that although our cycle-aggregative self-supervised strategy improves 3D representations in novel views, there are still artifacts near edges with significant viewpoint changes.
To address this, we introduce a fixed video in-painting model~\cite{zhou2023propainter} in the second training stage for artifact localization and additional refinement. 

\noindent\textbf{Artifacts localization using normal map.}
When rendering novel view images from $\hat{GS_0}$, we also render the alpha map and calculate the normal map from the depth. 
We found that alpha and normal maps effectively help identify artifacts. Specifically, if the alpha value is low and the angle between the normal and viewing direction is small, it is likely an artifact. This can be formalized as follows: 
\begin{equation}
\setlength{\abovedisplayskip}{6pt}
\setlength{\belowdisplayskip}{6pt}
    M_{\text{artifact}} = \left( \hat{A} < \tau \right) \land \left( \cos^{-1}\left( \hat{N} \cdot \mathbf{v} \right) < \tau_{\theta} \right)
    \label{equ:artifact}
\end{equation}
where $\hat{N}$ is the normal map calculated from depth map, $\mathbf{v}$ is the viewing direction, and  $\tau$ and $\tau_{\theta}$ are small thresholds. 

The rationale for using the normal map to detect artifacts is that poorly learned GS primitives in novel views tend to produce surface normals nearly parallel to the viewing direction. As shown in Fig.~\ref{fig:normal}, we use the normal map to identify these artifacts.
For small novel view changes, the rendered image and normals remain reasonable, and the corresponding mask is mostly zero. However, with larger view changes, holes appear near the frog's eyes due to a lack of Gaussian primitives, as indicated by very low alpha values. Additionally, redundant artifacts emerge on the top of the frog's head, visible in the rendered normal map. 
The mask on the right effectively highlights these artifact locations, highlighted by red arrows and boxes. 

\begin{figure*}[!t]
    \centering
    \includegraphics[width=0.95\linewidth]{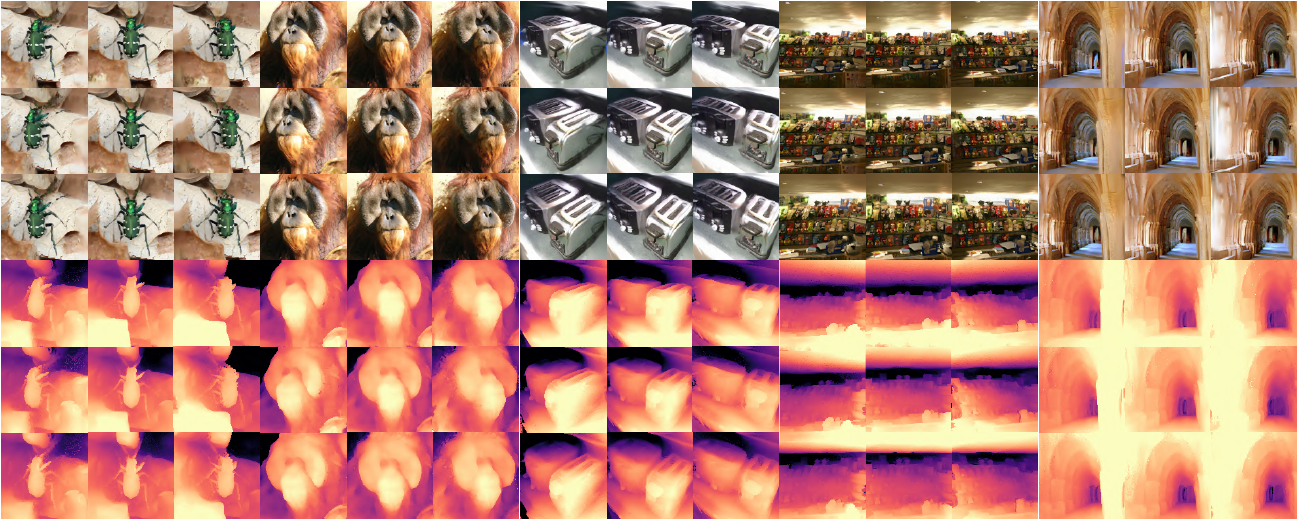}
    \vspace{-10pt}
    \caption{\textbf{Qualitative visualization of rendered images and depth maps on the ImageNet dataset.} Our method can generate novel view images along with corresponding depth maps for input images across various categories. }
    \label{fig:imagenet}
    \vspace{-18pt}
\end{figure*}

\par\noindent\textbf{Online sequence in-painting.}
After localizing the artifacts, we uniformly sample 16 views between $\text{view}_0$ and $\text{view}_1$, and render both the images $\hat{I^m}$ and depth maps $\hat{D^m}$ for these views from $\hat{\text{GS}}_0$. We then apply the video in-painting model~\cite{zhou2023propainter}, guided by masks generated from the alpha and normal maps, to fill in the missing regions in these 16 frames. The in-painted textures are subsequently used to supervise the rendered images in sequence. 
The in-painting process can be formulated as:
\begin{equation}
\setlength{\abovedisplayskip}{6pt}
\setlength{\belowdisplayskip}{6pt}
\left( I_{\text{in}}^{1}, \dots, I_{\text{in}}^{16} \right) = \mathcal{F}_{\text{in}} \left( \left( \hat{I^1}, M^1 \right), \dots, \left( \hat{I^{16}}, M^{16} \right) \right)
\end{equation}
where $I_{\text{in}}^m$ represents the in-painted image for the $m$-th frame, $\hat{I^m}$ is the rendered image, and $M^m$ is the mask generated from Equ.~\ref{equ:artifact}. Fig.~\ref{fig:normal} (b) shows one view of the in-painted results. Then, the corresponding video in-painting loss is defined as:
\begin{equation}
\setlength{\abovedisplayskip}{-2pt}
\setlength{\belowdisplayskip}{8pt}
L_{\text{video}} = \sum_{m=1}^{16} \left\lVert M^m \cdot ( \hat{I}^m - I_{in}^m ) \right\rVert_1
\end{equation} 
The in-painting process is performed online, and this video loss $L_{\text{video}}$ is added to $L_{\text{total}}$ to fine-tune the pre-trained model with a smaller learning rate. 
\section{Experiments}
\label{sec:exp}

\subsection{Experimental Setup}
\par\noindent\textbf{Dataset.}
We conducted experiments on ImageNet~\cite{deng2009imagenet}, Dogs~\cite{mokady2022self}, SDIP Elephants~\cite{mokady2022self}, and LSUN Horses~\cite{yu2015lsun}, following previous methods. ImageNet contains 1000 categories, while the others are single-class images. For fair comparison, we use pseudo-ground-truth depth from LeReS~\cite{Wei2021CVPR}, though our approach supports other depth estimators. Following G3DR~\cite{reddy2024g3dr}, we train on a filtered subset (420K images) and evaluate on the full set (1.2M images). Most of our experiments are conducted on ImageNet. 

\par\noindent\textbf{Metrics.}
For the generated images, we compute Fréchet Inception Distance (FID)~\cite{heusel2017gans} and Inception Score (IS) on the full ImageNet training set. For the predicted depth maps, 
we compute the Non-Flatness Score (NFS)~\cite{3dgp}, a no-reference depth quality metric that evaluates the continuity by analyzing the histogram distribution of the depth map. 

\par\noindent\textbf{Baseline.}
Since our model does not rely on conditional image generation, we primarily compare it with methods in ``3D-aware generation model with RGBD input" component of the state-of-the-art two-stage methods: $\text{IVID}_{\text{ICCV'23}}$~\cite{xiang2023ivid} and $\text{G3DR}_{\text{CVPR'24}}$~\cite{reddy2024g3dr}. 

\par\noindent\textbf{Settings.} 
We evaluate our model with two different input types and compare it against previous methods. First, we use ImageNet images as input to reconstruct the original view for comparison with G3DR. Second, we used 10k images sampled by the IVID~\cite{xiang2023ivid} to render novel views, to verify the effectiveness of the entire two-stage pipeline with the same 2D conditional image generation module. Yaw and pitch angles of novel camera are sampled from Gaussian distributions with variances of 0.3 and 0.15, respectively.

\subsection{Main Results}
\par\noindent\textbf{Quantitative results.}
Fig.~\ref{fig:imagenet}, Tab.~\ref{tab:imagenet_fidelity}, and Tab.~\ref{tab:imagenet_fidelity_128} present our qualitative and quantitative results on ImageNet. 
In Tab.~\ref{tab:imagenet_fidelity}, we compare G3DR's ``single-view 3D-aware model" $f_\textit{trigen}$ with ours, using the full ImageNet as input. The outputs are then upscaled to $256^2$ via Real-ESRGAN~\cite{wang2021realesrgan} for comparison. 
The top part of Tab.~\ref{tab:imagenet_fidelity} shows our method significantly outperforms G3DR in image fidelity and depth consistency. 
To eliminate the potential gap introduced by Real-ESRGAN, we also compare our $128^2$ model with $f_\textit{trigen}$ using metrics directly at 128 resolution. 
Tab.~\ref{tab:imagenet_fidelity} confirms that our method outperforms G3DR at both resolutions.

We further adopt another setting to evaluate the quality of generated novel views images by ``2D conditional generation + 3D generation". 
In Tab.~\ref{tab:imagenet_fidelity_128}, we use 10K sampled images from IVID-stage1 with class labels as input. 
Tab.~\ref{tab:imagenet_fidelity_128} shows that our method achieves the best FID/NFS and competitive IS, highlighting the advantage of pixel-aligned GS in 3D-aware generation. 

We provide additional quantitative evaluations on LSUN Horses~\cite{yu2015lsun}, SDIP Dogs, and Elephants~\cite{mokady2022self}, summarized in Tab.~\ref{tab:dogs} at 256 resolution. Our method outperforms G3DR~\cite{reddy2024g3dr} with better FID and NFS, indicating that our method generates more realistic images with smoother depth continuity. 
Since G3DR's pre-trained models for these datasets are unavailable, we train the $f_\textit{trigen}$ models using the released code, and upsample G3DR~\cite{reddy2024g3dr}'s $128^2$ outputs to $256^2$ with Real-ESRGAN~\cite{wang2021realesrgan}.

\begin{table}[!t]
\centering
\resizebox{0.48\textwidth}{!}{
\begin{tabular}{llll} 
\toprule
Method & FID $\downarrow$ & IS $\uparrow$ & NFS $\uparrow$ \\ 
\midrule
G3DR-$f_\textit{trigen}$ + RealESRGAN ($256^2$) & 24.9 & 78.2 & 34.1 \\ 
\textbf{Our F3D-Gaus} ($256^2$) & \textbf{1.6} & \textbf{308.6} & \textbf{40.4} \\ 
\midrule
\rule{0pt}{10pt}G3DR-$f_\textit{trigen}$ ($128^2$) & 14.5 & 86.3 & 34.0 \\ 
\textbf{Our F3D-Gaus} ($128^2$) & \textbf{1.2} & \textbf{202.8} & \textbf{40.6} \\ 
\bottomrule
\end{tabular}
}
\vspace{-8pt}
\caption{\textbf{Comparison on ImageNet with G3DR.} Our \textbf{F3D-Gaus} surpasses the state-of-the-art 3D generation methods G3DR~\cite{reddy2024g3dr} in FID, IS, and NFS metrics in both $128^2$ and $256^2$ resolution. }
\label{tab:imagenet_fidelity}
\vspace{-8pt}
\end{table}

\begin{table}[!t]
\centering
\resizebox{0.38\textwidth}{!}{
\begin{tabular}{llll} 
\toprule
Method            & FID $\downarrow$    & IS $\uparrow$  & NFS $\uparrow$   \\ \midrule
IVID~\cite{xiang2023ivid} 128x                    & 16.5                & \textbf{216.4}  & 25.7        \\ 
G3DR~\cite{reddy2024g3dr} 128x                    & \underline{15.8}                & 135.3 & \underline{36.1}           \\ \midrule
\textbf{Our F3D-Gaus} 128x                             & \textbf{15.5}        & \underline{182.7} & \textbf{37.4}  \\  
\bottomrule 
\end{tabular}
}
\vspace{-8pt}
\caption{\textbf{Comparison with state-of-the-art methods on 10k sampled images.} \textbf{F3D-Gaus} outperforms in two of three key metrics, highlighting its strengths in fidelity and geometry consistency. }
\label{tab:imagenet_fidelity_128}
\vspace{-8pt}
\end{table}

\begin{table}[!t]
\centering
\resizebox{0.37\textwidth}{!}{
\begin{tabular}{lll} 
\toprule
Method                             & Training $\downarrow$   & Inference (s) $\downarrow$    \\ \midrule 
IVID  \cite{xiang2023ivid}         & N/A                          & 20+128              \\ 
G3DR~\cite{reddy2024g3dr}          & 14.5                          & 0.9+3.6              \\ \midrule 
\textbf{Our F3D-Gaus}              & \textbf{13.0}                     & \textbf{0.8+1.1}         \\      
\bottomrule 
\end{tabular}
}
\vspace{-8pt}
\caption{\textbf{Efficiency comparison with state-of-the-art methods.} Our \textbf{F3D-Gaus} achieves the fastest training (measured in A100 days) and inference speed. }
\label{tab:inference}
\vspace{-20pt}
\end{table}

\par
\begin{figure}[!t]
    \centering
    \includegraphics[width=\linewidth]{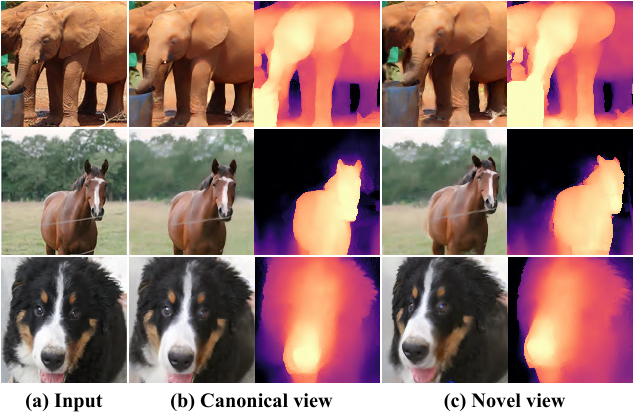}
    \vspace{-20pt}
    \caption{\textbf{Qualitative results of rendered images and depth maps} on SDIP Elephants~\cite{mokady2022self}, LSUN Horses~\cite{yu2015lsun}, and Dogs~\cite{mokady2022self}.}
    \label{fig:animal}
    \vspace{-18pt}
\end{figure}

\noindent\textbf{Qualitative results.} 
Fig.~\ref{fig:imagenet} showcases 3DGS renderings from our model in both canonical and novel views across a wide range of categories, including slender-legged insects, animals, household items, detailed objects, and large structures. These results highlight our model’s ability to handle diverse data and generate high-quality outputs. 
Fig.~\ref{fig:animal} provides additional qualitative results on three single-class datasets, further illustrating the effectiveness of our approach. Beyond accurately reconstructing the original view, our method generates plausible and coherent novel views.

\par
\begin{figure}[!tp]
    \centering
    \includegraphics[width=0.98\linewidth]{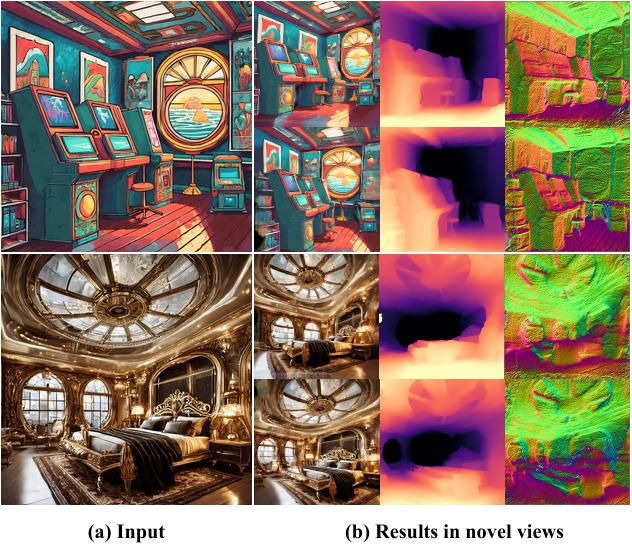}
    \vspace{-8pt}
    \caption{\textbf{Results on two complex indoor scene images outside the ImageNet dataset.} The rendered novel view images, along with their corresponding depth and normal maps, effectively demonstrate the generalization capability of our model. }
    \label{fig:abl_ood}
    \vspace{-20pt}
\end{figure}

\noindent\textbf{Out-of-domain samples.} 
Fig.~\ref{fig:abl_ood} shows the results of our method on more complex, scene-level inputs that fall outside the scope of ImageNet. 
The out-of-domain samples, including intricate indoor scenes and multi-object environments, introduce significantly higher complexity than ImageNet's object-centric images. 
Despite the challenge, our model trained on ImageNet at $256^2$ resolution successfully produces high-quality novel view renderings, including images, depth, and normal maps across various viewpoints, demonstrating its impressive generalization capabilities.

\par\noindent\textbf{Efficiency.} 
Tab.~\ref{tab:inference} compares our method's efficiency with IVID and G3DR. 
Our approach reduces the training time from 14.5 to 13 A100 days. Inference time is reported as "Time to obtain the 3D representation + Time to render 128 novel views." Leveraging pixel-aligned GS, our method achieves the fastest runtime. 
While G3DR's training and inference times are close to ours, it only supports $128^2$ inputs, which requires an additional super-resolution step to upscale $128^2$ outputs to $256^2$. Our pixel-aligned Gaussian design enables direct training and testing at $256^2$, eliminating post-processing overhead not reflected in Tab.~\ref{tab:inference}.

\begin{table}[!t]
\resizebox{0.47\textwidth}{!}{
\centering
\begin{tabular}{lllllll}
\toprule
\multirow{2}{*}{Method ($256^2$)}  &  \multicolumn{2}{c}{Dogs} & \multicolumn{2}{c}{Horses} & \multicolumn{2}{c}{Elephants}\\
  & FID$\downarrow$ & NFS $\uparrow$   & FID$\downarrow$ &NFS $\uparrow$ & FID$\downarrow$ & NFS $\uparrow$\\ \midrule
\multirow{2}{*}{\parbox{2.4cm}{G3DR-$f_\textit{trigen}$\\+ RealESRGAN}} & \multirow{2}{*}{14.11} & \multirow{2}{*}{37.16} & \multirow{2}{*}{11.65}  & \multirow{2}{*}{32.31} & \multirow{2}{*}{24.39} & \multirow{2}{*}{33.02}  \\ \\  \midrule
\textbf{Our F3D-Gaus }  & \textbf{2.53}  & \textbf{41.22}  & \textbf{1.14} & \textbf{39.67}  & \textbf{2.96}  & \textbf{38.92}       \\
\bottomrule 
\end{tabular}
}
\vspace{-8pt}
\caption{\textbf{Quantitative evaluation} on the LSUN Horses~\cite{yu2015lsun}, SDIP Dogs, and Elephants~\cite{mokady2022self} datasets. }
\vspace{-12pt}
\label{tab:dogs}
\end{table}

\begin{figure*}[!tp]
    \centering
    \includegraphics[width=\linewidth]{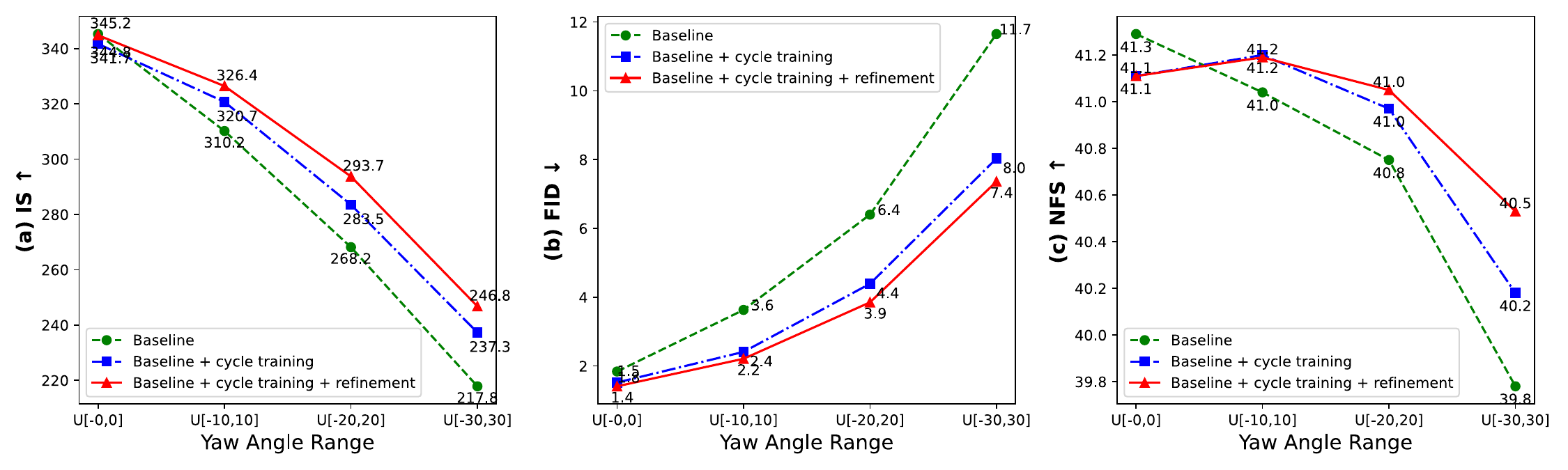}
    \vspace{-26pt}
    \caption{\textbf{Performance comparison of three $128^2$ models across varying yaw angle ranges.} The FID, IS, and NFS metrics are evaluated on the filtered subset of ImageNet. The x-axis represents the yaw angle ranges, while the y-axis denotes the corresponding metric values. }
    \label{fig:abl_all}
    \vspace{-18pt}
\end{figure*}

\subsection{Ablation Study} 
\label{sec:abl}
\begin{figure}[!tp]
    \centering
    \includegraphics[width=0.98\linewidth]{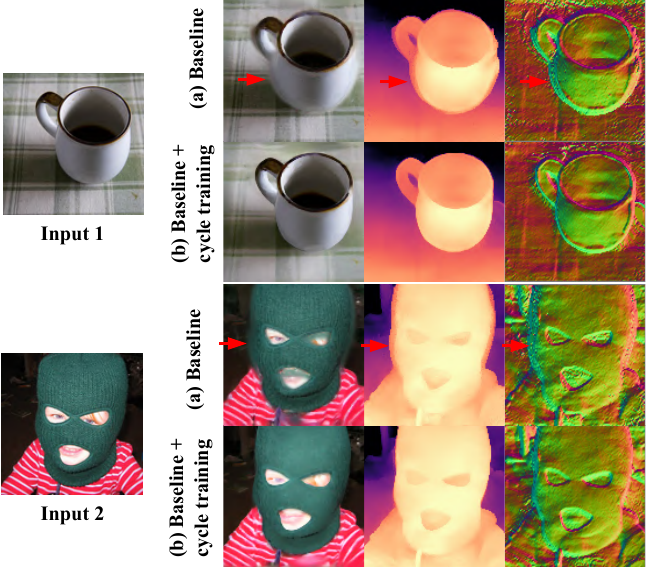}
    \vspace{-8pt}
    \caption{\textbf{Demonstration of the effectiveness of cycle-aggregative self-supervised training.} The comparison between (a) and (b) clearly highlights that our cycle-aggregative training strategy significantly mitigates ghosting artifacts. }
    \label{fig:abl_cycle}
    \vspace{-17pt}
\end{figure}

Fig.~\ref{fig:abl_all} and Fig.~\ref{fig:abl_cycle} demonstrate the effectiveness of our proposed modules quantitatively and qualitatively. 

\par\noindent\textbf{Overall analysis.}
Fig.~\ref{fig:abl_all} shows how different components of our method affect various metrics as the view angle increases. For efficiency, these metrics are computed on a ImageNet subset at 128 resolution. 
The three subplots in Fig.~\ref{fig:abl_all} correspond to IS, FID, and NFS, each with three curves representing different variants. 
The green line represents the baseline, using only the pixel-aligned representation, without cycle-aggregative training or second-stage refinement. 
In this variant, no complementary aggregation is applied, $\mathcal{L}_{\text{cycle}}$ is removed from Equ.~\ref{equ:total}, and $\mathcal{L}_{\text{novel}}$ is applied to $\tilde{I_1}$ and $\tilde{D_1}$, rather than $\hat{I_1}$ and $\hat{D_1}$. 
The blue line represents the model with cycle-aggregative training, while the red line represents the full model, incorporating both cycle-aggregative training and second-stage refinement.

\par\noindent\textbf{Pixel-aligned 3DGS.} 
Fig.~\ref{fig:abl_all} shows that at 0° angle, all three variants perform similarly, with the baseline (green line) even achieving the highest IS and NFS. This suggests that the generalized 3DGS representation alone can produce high-quality frontal view reconstructions. 
However, as the yaw angle increases, the baseline's performance degrades sharply. For angles sampled from [-30°, 30°], its IS drops to 217.8, FID rises to 11.7, 
indicating that it struggles to preserve texture and geometry effectively in novel views.

\par\noindent\textbf{Cycle-aggregative training.} 
In contrast, the blue line, which incorporates cycle-aggregative training, performs better at wider viewing angles. 
For instance, at [-30°, 30°] yaw angle, its FID drops from 11.7 to 8.0 (Fig.~\ref{fig:abl_all} (b)), showing that cycle-aggregative training improves 3D perception using monocular datasets.
Fig.~\ref{fig:abl_cycle} further illustrates its effect on novel views. 
The right side shows rendered images, depth, and normal maps for two input images. 
Without cycle training, novel views exhibit noticeable ghosting artifacts, \textit{e.g.,} input 1's mug has a faint extra edge, and input 2's head appears as multiple overlapping layers. 
These artifacts indicate that 3DGS representations derived from different viewpoints are not well-aligned, which contradicts our goal of learning a consistent 3D-aware representation. 
However, after applying cycle-aggregative training, the rendered results become much more coherent, eliminating ghosting artifacts. This demonstrates its effectiveness in producing consistent and aligned 3D representations across views. 

\par\noindent\textbf{Geometry-guided refinement. } 
In Fig.~\ref{fig:abl_all}, comparing the blue and red lines, we observe that the second-stage geometry-guided refinement obviously improves performance. The red line achieves the best results among all three metrics, indicating that this refinement further enhances both image and geometry rendering.

\section{Conclusion}
\label{sec:conclusion}

We are the first to apply pixel-aligned Gaussian Splatting representations to generalizable 3D-aware generation on monocular datasets. 
In the absence of multi-view or dynamic data, we propose a self-supervised cycle training strategy that effectively merges multiple geometry-aligned 3D representations for better 3D-aware capability. 
We also incorporate video-based priors for geometry-aware detail refinement, particularly in wide-angle views. 
Experimental results demonstrate that our method outperforms previous baselines in both effectiveness and efficiency.


\setcounter{page}{1}
\maketitlesupplementary

\appendix
In this supplementary material, we provide more additional experiments In this supplementary material, we provide more additional experiments in Sec.~\ref{sec:supple_sec_1}. We also present a video demo for more qualitative results as shown in Sec.~\ref{sec:supple_video}.

\section{Additional Experiments}\label{sec:supple_sec_1}
\subsection{More Qualitative Results}\label{sec:a3}
\begin{figure}[!t]
    \centering
    \includegraphics[width=\linewidth]{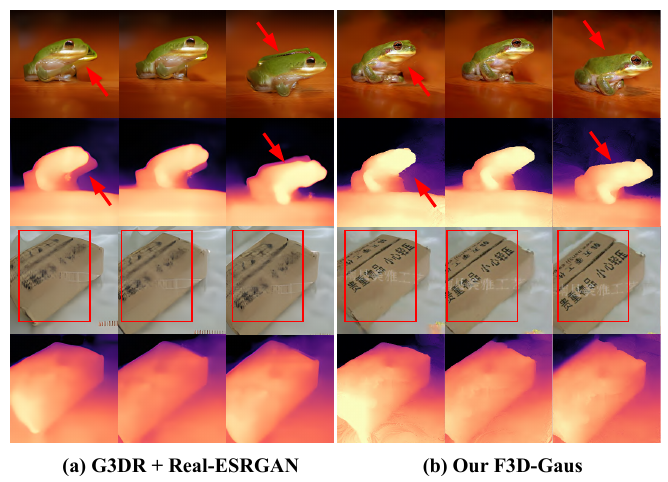}
    \vspace{-16pt}
    \caption{\textbf{Qualitative comparison with G3DR~\cite{reddy2024g3dr}. } We use Real-ESRGAN\cite{wang2021realesrgan} to upsample G3DR's $128^2$ outputs to $256^2$. For our F3D-Gaus, we accept $256^2$ resolution inputs and render directly at the same $256^2$ resolution, without requiring post-processing. }
    \label{fig:g3dr}
    \vspace{-6pt}
\end{figure}

\begin{figure*}[!tp]
    \centering
    \includegraphics[width=\linewidth]{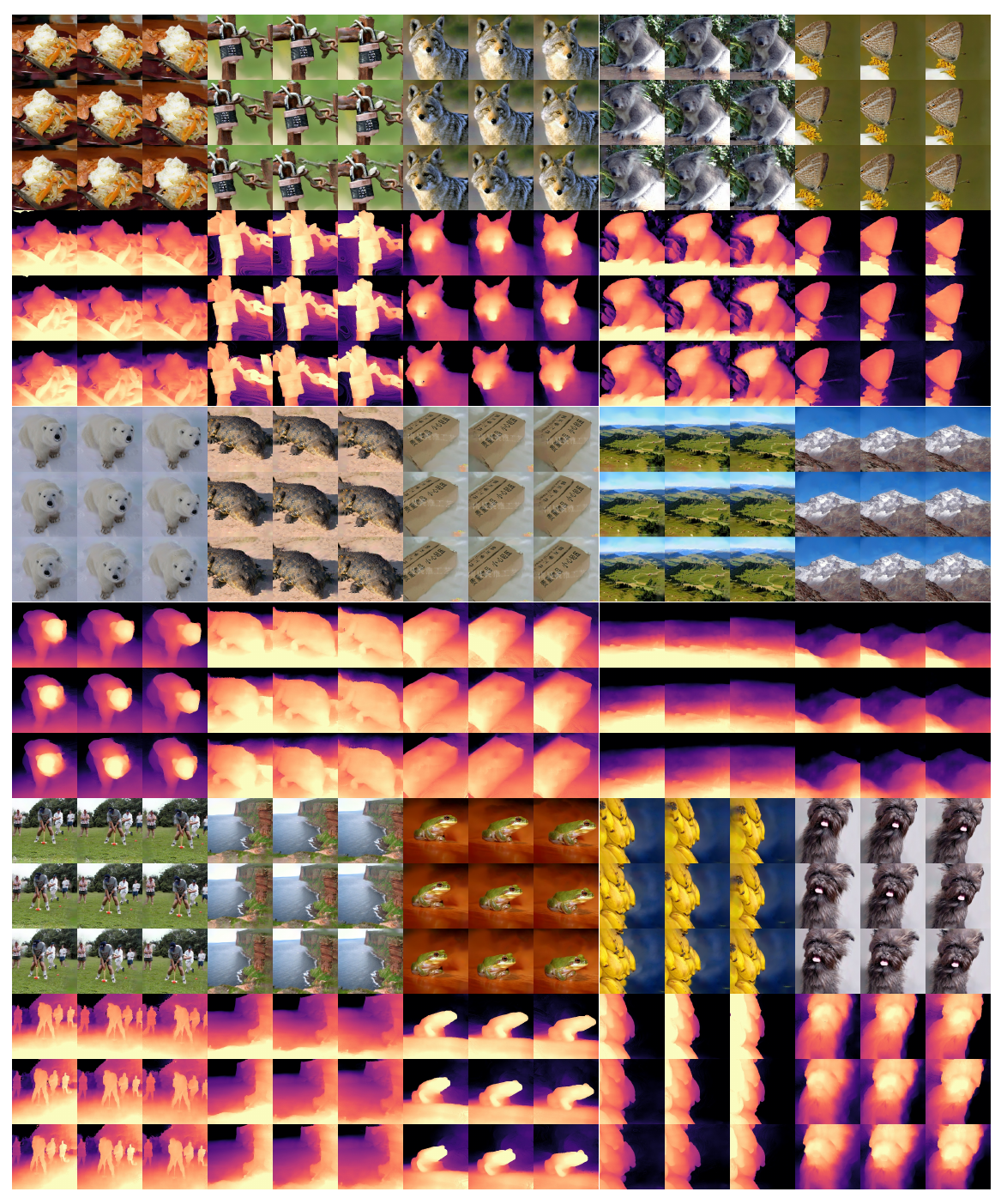}
    \vspace{-16pt}
    \caption{\textbf{Additional qualitative visualization of rendered images and depth maps on the ImageNet dataset. } }
    \label{fig:imagenet1}
\end{figure*}

\begin{figure*}[!tp]
    \centering
    \includegraphics[width=\linewidth]{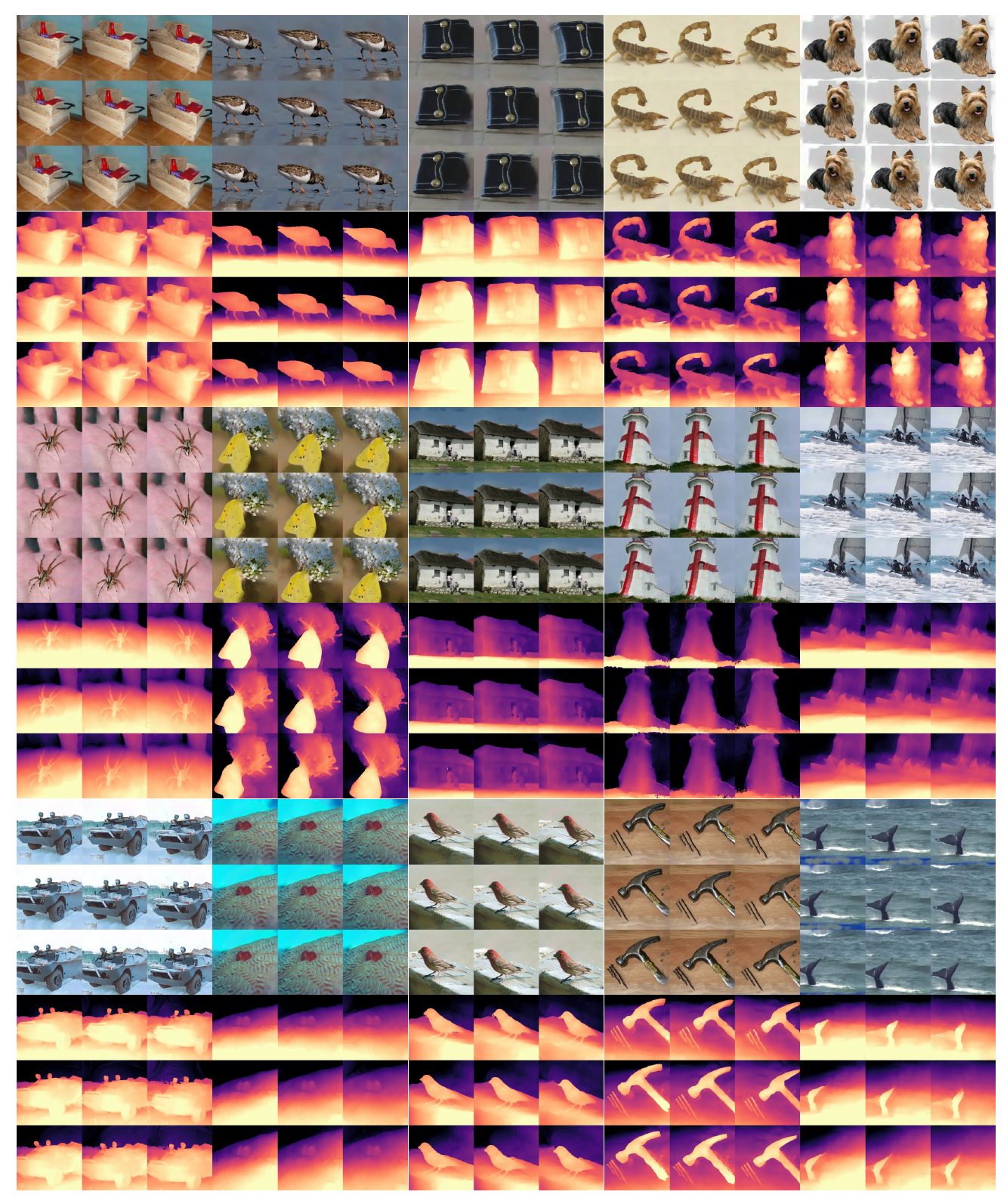}
    \vspace{-16pt}
    \caption{\textbf{Another qualitative visualization of rendered images and depth maps on the ImageNet dataset. } }
    \label{fig:imagenet2}
\end{figure*}

Fig.~\ref{fig:g3dr} presents a qualitative comparison between our method and G3DR. Due to the absence of the upsampling code of G3DR~\cite{reddy2024g3dr}, we employ Real-ESRGAN~\cite{wang2021realesrgan}, a super-resolution model known for its strong generalization on diverse images, to upsample G3DR's $128^2$ outputs to $256^2$. Although the super-resolution model can enhance sharpness in low-resolution results, it fails to address G3DR's inherent issues, such as multi-head artifacts and lack of fine details. The red arrows and boxes in Fig.~\ref{fig:g3dr} highlight that our approach achieves better multi-view consistency and preserves more intricate details. 
In Fig.~\ref{fig:imagenet1} and Fig.~\ref{fig:imagenet2}, we provide additional novel view synthesis results on the ImageNet dataset at the resolution of $256^2$. These qualitative results highlight that our method consistently produces images and depth across a diverse range of categories.

\begin{figure}[!t]
    \centering
    \includegraphics[width=\linewidth]{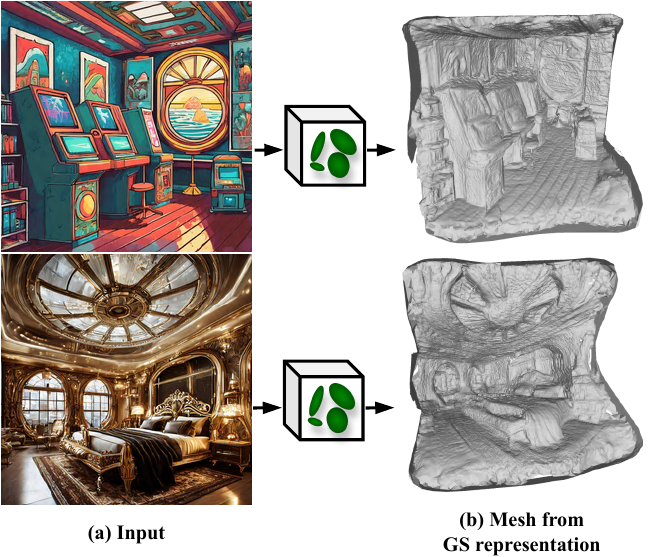}
    \vspace{-15pt}
    \caption{\textbf{Visualization of the mesh extracted from the predicted 3DGS. } We borrow the pipeline of GOF~\cite{Yu2024GOF} for the mesh extraction. The mesh is directly derived from the 3DGS representation and does not rely on image-based optimization methods. }
    \label{fig:mesh}
    \vspace{-13pt}
\end{figure}

\subsection{Mesh Extraction}
Fig.~\ref{fig:mesh} shows the results of mesh extraction from the predicted 3DGS. Our 3DGS representation is based on GOF~\cite{Yu2024GOF} for accurate depth and normal rendering, which naturally supports surface reconstruction. After we get the predicted 3DGS from a single image, we extract meshes using GOF's tetrahedral grid generation combined with a binary search. Fig.~\ref{fig:mesh} demonstrates the capability of our method to predict mesh from a single image via a geometry-driven pipeline leveraging 3DGS. Please note that the mesh is directly derived from the 3DGS representation and does not rely on image-based optimization methods. 

\subsection{Ablation on Depth Estimation}
Our method currently relies on depth input, but it is technically feasible to modify the U-Net backbone to predict the depth map simultaneously. This modification would enable our method to function solely with image input while maintaining its existing capabilities. To explore this, we experimented with extending the U-Net to output an additional channel for pseudo-depth map regression, while optimizing the total loss described in the main text. The quantitative results are shown in Tab.~\ref{tab:predd}. 
The results indicate that when the U-Net is tasked with predicting the depth map, both image metrics (FID, IS) and depth metrics (Depth Accuracy, NFS) degrade. Although the FID and IS scores remain strong compared to 3D-aware methods, the depth accuracy significantly declines. This suggests that the regressed depth map is inaccurate, and incorporating depth prediction as an additional optimization objective disrupts the learning process for image generation. 
We attribute this performance decline to the limitations of the current U-Net architecture, which is not specifically designed for depth map regression. Additionally, balancing multiple optimization objectives may require more sophisticated strategies. Reducing reliance on depth input remains a key focus for our future work. 

\begin{table}[!t]
\centering
\resizebox{0.5\textwidth}{!}{
\begin{tabular}{lllll} 
\toprule  
Variants (128x)     
& FID $\downarrow$ & IS $\uparrow$ & Depth Acc $\downarrow$ & NFS$\uparrow$ \\ \midrule
F3D-Gaus                             & \textbf{1.2}  & \textbf{202.8}  & \textbf{0.16} & \textbf{40.5}             \\ 
F3D-Gaus w/ predicted depth          & 1.7  & 195.0  & 0.33 & 36.6             \\  
\bottomrule 
\end{tabular}
}
\vspace{-4pt}
\caption{\textbf{Comparison experiment with predicting depth simultaneously. }When the U-Net is tasked with predicting the depth map simultaneously, both image metrics (FID, IS) and depth metrics (Depth Accuracy, NFS) degrade.}
\vspace{-10pt}
\label{tab:predd}
\end{table}

\subsection{Why no conditional generation model? }
\begin{figure}[htbp]
    \centering
    \vspace{-13pt}
    \includegraphics[width=1.0\linewidth]{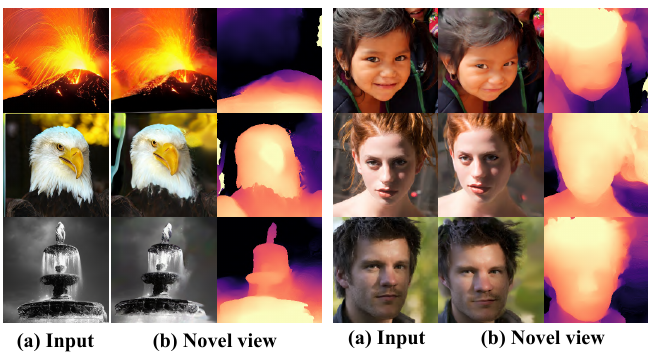}
    \vspace{-10pt}
    \caption{\textbf{Qualitative results on VAR~\cite{tian2024visual} output (left) and FFHQ images (right).}}
    \label{fig:var}
    \vspace{-2pt}
\end{figure}

The pipeline of state-of-the-art methods G3DR~\cite{reddy2024g3dr} and IVID~\cite{xiang2023ivid} can be divided into a ``class-conditioned RGBD generation model'' and a ``3D-aware generation model with RGBD input'', where we focus on the latter. 
Actually, training an additional conditional generation model to fit the ImageNet distribution is straightforward with existing frameworks like GANs, Diffusion models, and VARs. Including a preceding RGBD generation (like G3DR and IVID) in evaluation introduces confounding factors that blend both models' performance, detracting from our core contribution. Although we did not train a conditional generation model, our method still shows extensibility to images generated by existing models within the ImageNet distribution. As shown in Fig.~\ref{fig:var}, our method performs well on in-domain generated images (from VAR~\cite{tian2024visual} trained on ImageNet) and even out-of-domain images (FFHQ~\cite{karras2019style}), demonstrating its generalization ability. 

\section{Video Demo}\label{sec:supple_video} 
As a supplement to Sec.~\ref{sec:a3}, we have included a video demo to showcase the quantitative results in a video format. As shown in Fig.\ref{fig:g3dr}, we use Real-ESRGAN~\cite{wang2021realesrgan} to upsample $128^2$ outputs of G3DR~\cite{reddy2024g3dr} to $256^2$.

\end{document}